\documentclass{article}

\usepackage[final,main]{neurips_2026}

\usepackage[utf8]{inputenc}
\usepackage[T1]{fontenc}
\usepackage{hyperref}
\usepackage{url}
\usepackage{booktabs}
\usepackage{amsfonts}
\usepackage{amsmath}
\usepackage{amssymb}
\usepackage{nicefrac}
\usepackage{microtype}
\usepackage[table]{xcolor}
\definecolor{best}{HTML}{7FCBAA}
\definecolor{secondbest}{HTML}{D5EFE2}
\usepackage{graphicx}
\usepackage{multirow}

\newcommand{\method}{Ours\xspace}
\providecommand{\xspace}{}

\title{Edit Where You Mean: Region-Aware Adapter Injection\\
       for Mask-Free Local Image Editing}

\author{%
  Honghao Cai$^{1,5,*}$ \quad
  Xiangyuan Wang$^{4,5,*}$ \quad
  Yunhao Bai$^{4,5}$ \quad
  Haohua Chen$^{2,5}$ \quad
  Tianze Zhou$^{3,5}$ \\
  Runqi Wang$^{5}$ \quad
  Wei Zhu$^{5}$ \quad
  Yibo Chen$^{5}$ \quad
  Xu Tang$^{5}$ \quad
  Yao Hu$^{5}$ \quad
  Zhen Li$^{1,\dagger}$ \\[0.4em]
  $^{1}$The Chinese University of Hong Kong, Shenzhen \quad
  $^{2}$Beijing University of Aeronautics and Astronautics \\
  $^{3}$Tsinghua University \quad
  $^{4}$Peking University \quad
  $^{5}$Xiaohongshu Inc. \\[0.3em]
  $^{*}$Equal contribution. \quad $^{\dagger}$Corresponding author.
}

\begin{document}

\maketitle

\begin{abstract}
Large diffusion transformers (DiTs) follow global editing instructions well but consistently leak \emph{local} edits into unrelated regions, because joint-attention architectures offer no explicit channel telling the network \emph{where} to apply the edit. We introduce \method, a co-trained, instruction- and region-aware adapter framework that retro-fits a frozen DiT into a precise local editor without modifying its backbone weights. A lightweight \emph{Block Adapter} at every transformer block injects a structured condition stream that factorizes \emph{what} to edit (instruction semantics) from \emph{where} to edit (spatial mask); a learned \emph{SpatialGate} routes the adapter signal selectively into the edit region while keeping the rest of the image near-identical to the source; and a \emph{Region-Aware Loss} focuses the training objective on the changing pixels. Because these components make the backbone's internal representation mask-aware end-to-end, a thin \emph{MaskPredictor} head trained jointly with the editor can ground the edit region directly from the instruction and source image --- eliminating any user-mask requirement at deployment. We evaluate on two complementary benchmarks: MagicBrush (paired ground-truth targets) to measure pixel-level preservation and edit accuracy, and Emu-Edit Test (no ground-truth images, 9 diverse edit categories) to stress-test instruction following and generalization across edit types. On both, \method achieves state-of-the-art results, simultaneously outperforming mask-free and oracle-mask baselines. A seven-variant ablation cleanly isolates the contribution of each component.
\end{abstract}

\section{Introduction}

\paragraph{Existing work and how they do it.}
Instruction-driven image editing has evolved through two main lines. The first, pioneered by InstructPix2Pix~\citep{instructpix2pix} and advanced by MagicBrush~\citep{magicbrush}, UltraEdit~\citep{ultraedit}, and the 2025--2026 DiT-class editors (BAGEL~\citep{bagel}, OmniGen2~\citep{omnigen2}, FLUX-Kontext~\citep{fluxkontext}, Step1X-Edit~\citep{step1xedit}, Emu-Edit~\citep{emuedit}), builds unified models that concatenate instruction tokens with image tokens and rely on \emph{joint attention} to fuse the two --- scaling from 1B UNets to multi-billion-parameter DiTs. The second line conditions editing on a user-provided binary mask: BrushNet~\citep{brushnet}, FLUX.1-Fill, and recent segment-then-edit pipelines pair an off-the-shelf SAM-class segmenter with an inpainter to restrict diffusion to a masked region. Some recent work~\citep{anyedit} attempts to unify both by jointly training edit and segmentation heads on a shared backbone.

\begin{figure}[t]
  \centering
  \includegraphics[width=\linewidth]{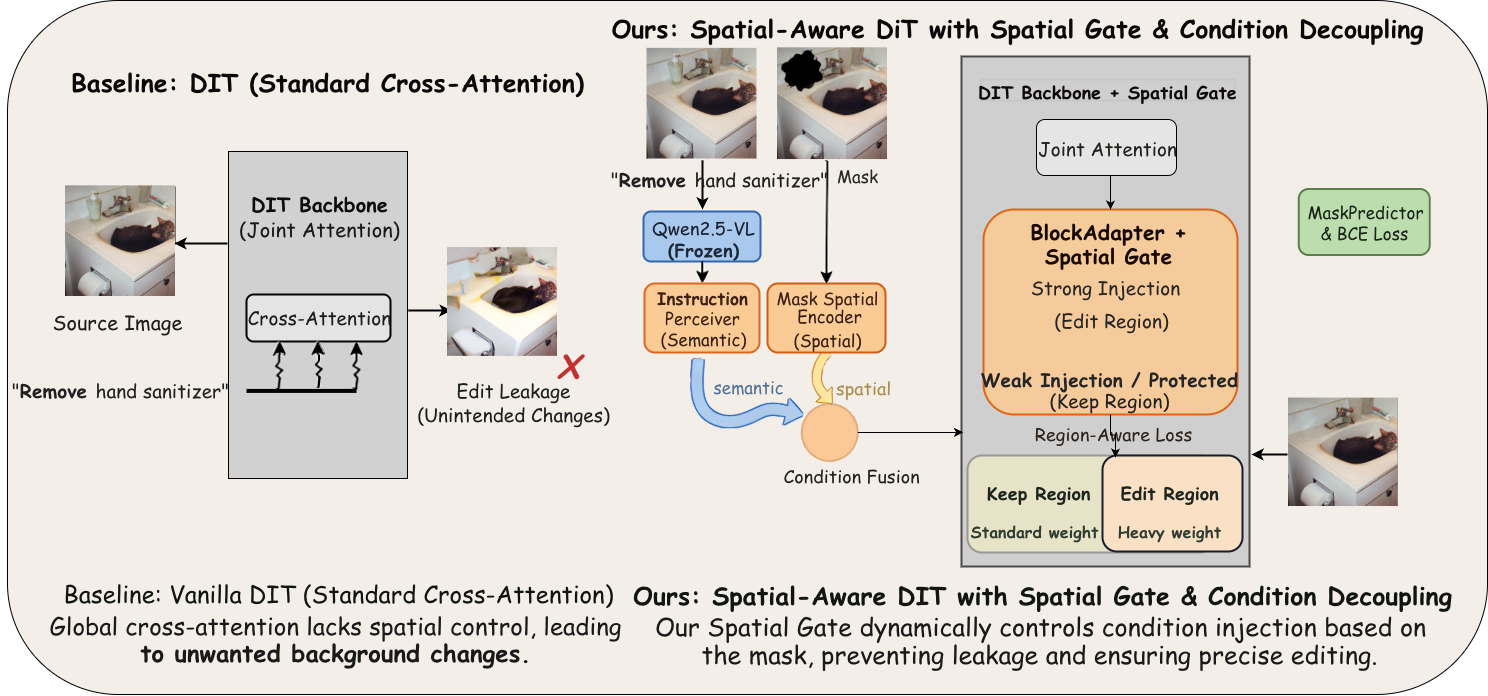}
  \caption{\textbf{Motivation.} \emph{Left:} a vanilla DiT leaks ``remove hand sanitizer'' into surrounding pixels. \emph{Right:} \method factorizes \emph{what} (instruction) from \emph{where} (mask), injects both via a per-block adapter modulated by a SpatialGate, and grounds the edit region automatically via a MaskPredictor --- no user mask needed at deployment.}
  \label{fig:teaser}
\end{figure}

\paragraph{Their problems.}
Both lines have fundamental limitations (see Fig.~\ref{fig:teaser}). Joint-attention editors offer no explicit channel telling the network \emph{where} to apply the edit: the instruction is fused as a flat token sequence with no spatial axis. As we show in \S\ref{sec:experiments}, this causes consistent leakage on local edits even at billions of parameters --- a request like ``change only the cup to red'' bleeds into the table, alters background lighting, or drifts unrelated geometry. Mask-conditioned editors restore spatial control but demand a pixel-accurate mask per edit, which is impractical for iterative workflows at scale; segment-then-edit pipelines add 200--500\,ms overhead and inherit the segmenter's failure modes on referring-expression edits. Jointly-trained edit+segment heads~\citep{anyedit} couple two losses on the same backbone, which we find leads to a self-reinforcing failure mode: the backbone lacks an explicit mask-aware internal representation for the segmentation head to read, so neither head can bootstrap the other effectively. Industrial use cases --- e-commerce product editing, ad creative iteration, photo retouching --- demand precisely local edits with zero collateral damage outside, a capability that no current approach reliably delivers without user intervention.

\paragraph{How we solve it.}
We observe that a frozen large-scale DiT can be retro-fitted into a precise local editor without modifying its backbone weights, by providing the structural primitives --- ``where'' and ``what'' as separate streams --- that joint attention currently lacks. We introduce \method, a co-trained, instruction- and region-aware adapter framework with five jointly optimized components: (1)~a \textbf{Block Adapter} at every DiT block that injects a structured condition stream via cross-attention with gated-residual fold-back (only adapters trainable, backbone frozen); (2)~a \textbf{Condition Encoder} that explicitly factorizes the edit mask into spatial tokens and compresses instruction encoder hidden states into semantic tokens via a Perceiver projector, fused by a small transformer; (3)~a \textbf{SpatialGate} producing per-token soft gates that route the adapter signal selectively into the edit region while keeping the rest near-identical to the source; (4)~a \textbf{Region-Aware Loss} that re-weights the velocity objective by the mask so gradients focus on changing pixels; and (5)~a thin \textbf{MaskPredictor} head (FiLM-conditioned cross-attention decoder with decoupled BCE+Dice loss) that grounds the edit region directly from the instruction and source image. Because the first four components make the backbone's internal representation mask-aware end-to-end during training, the MaskPredictor can read out this representation to accurately localize edits --- a structural consequence of how the editor was trained. The deployed system requires only the source image and a natural-language instruction, with no user mask at any point.

\paragraph{Contributions.}
We contribute (i)~a co-trained, instruction- and region-aware adapter framework for a frozen DiT --- Block Adapter, Condition Encoder, SpatialGate, and Region-Aware Loss act jointly to make the backbone's internal representation mask-aware end-to-end (${\sim}25\%$ parameter overhead, backbone gradients zero); (ii)~a thin MaskPredictor head (decoupled BCE+Dice loss) that reads this mask-aware representation to ground the edit region from instruction and source image alone, requiring no external segmenter; and (iii)~state-of-the-art edit fidelity and accurate grounding from a single forward pass with no user mask --- achieving SOTA on both MagicBrush~(pixel-level preservation) and Emu-Edit Test~(instruction following across 9 edit categories), simultaneously outperforming mask-free and oracle-mask baselines.

\section{Related Work}
\label{sec:related}

\paragraph{Instruction-guided image editing.}
Editing-from-instruction was popularized by InstructPix2Pix~\citep{instructpix2pix} which fine-tunes a diffusion model on synthetic (source, instruction, target) triplets and established the now-standard pipeline of full-backbone fine-tuning. MagicBrush~\citep{magicbrush} contributed a manually-annotated dataset and a stronger baseline that for two years held the de-facto leaderboard on its own dev split. MGIE~\citep{mgie} couples a multimodal LLM to expand short instructions into rich latent guidance, while HQ-Edit~\citep{hqedit}, UltraEdit~\citep{ultraedit}, SmartEdit~\citep{smartedit} and SEED-Data-Edit~\citep{seededit} progressively scale data and models within the SD-1.5 family. The 2025--2026 wave of DiT-class editors --- Emu-Edit~\citep{emuedit}, AnyEdit~\citep{anyedit}, Step1X-Edit~\citep{step1xedit}, FLUX.1-Kontext~\citep{fluxkontext}, BAGEL~\citep{bagel}, OmniGen2~\citep{omnigen2} and the proprietary Qwen-Image-Edit~\citep{qwen-image-2025} which we adopt as the frozen backbone --- has lifted absolute quality dramatically by scaling the backbone (7B+) and switching from cross-attention to \emph{joint attention} over concatenated $[\text{image};\text{text}]$ tokens. They share a common architectural property, however: the instruction is fused as a flat token sequence with no explicit spatial axis. As we quantify in \S\ref{sec:experiments}, this implicit-routing design leaks consistently on local edits even at scale. \method departs from this trend in two ways: (i) it keeps the backbone fully frozen and (ii) it injects spatial information through a small adapter that exposes ``where'' as a first-class channel.

\paragraph{Other recent open-source editors.}
The empirical comparison in Section~\ref{sec:experiments} additionally includes nine concurrent or recent systems. \textbf{ACE}~\citep{ace}, \textbf{PixWizard}~\citep{pixwizard} and \textbf{UniReal}~\citep{unireal} are full-fine-tune SDXL-class generalists that unify multiple edit tasks under a single instruction-following model. \textbf{VAREdit}~\citep{varedit} replaces the diffusion denoiser with a next-scale visual auto-regressive editor, while \textbf{EditMGT}~\citep{editmgt} adapts a masked generative transformer to the same protocol. On the same Qwen-Image-Edit backbone we adopt frozen, \textbf{UniWorld}~\citep{uniworld} extends the model with a unified vision-language formulation, \textbf{ReasonEdit}~\citep{reasonedit} augments inference with explicit thinking and reflection traces, and \textbf{ICEdit}~\citep{icedit} pursues mask-free editing through concise in-context visual prompts; \textbf{VIBE}~\citep{vibe} is a strong recent open editor included for completeness on the GEdit-Bench leaderboard. All nine either fully fine-tune the backbone or attach a heavy parallel branch; none exposes ``where'' as an explicit, learned, in-block spatial channel --- the design property of \method that drives the headline preservation gains in Section~\ref{sec:experiments}.

\paragraph{Mask-conditioned and mask-localized editing.}
A separate line conditions on user-provided binary masks~\citep{glide,blendedlatent,brushnet}: BrushNet adds a ControlNet-style branch that consumes the mask, while inpainting models~\citep{rombach2022stablediffusion} and the recent FLUX.1-Fill family at the 12B DiT scale only diffuse inside the mask. 2026 segmentation-then-edit pipelines~\citep{step1xedit,brushnet} pair an off-the-shelf SAM-class model with such an inpainter, paying a 200--500\,ms front-end overhead and inheriting the segmenter's failure modes on referring-expression edits. Mask-free alternatives include training-free attention-extraction methods~\citep{prompt2prompt,plugandplay,pix2pixzero} and jointly-trained edit+segment heads~\citep{anyedit} --- the latter coupling the two losses, which we find leads to a self-reinforcing failure mode (\S\ref{sec:maskpred}). \method differs on three axes: (i) the mask is produced by an in-network decoupled MaskPredictor in a single forward pass, with no external segmenter call; (ii) the mask is consumed not as a hard boundary but as the input to a learned soft SpatialGate (\S\ref{sec:spatialgate}), avoiding seam artefacts; (iii) the deployed inference path needs no user mask at any point.

\paragraph{Parameter-efficient adaptation of large diffusion models.}
LoRA~\citep{lora}, IP-Adapter~\citep{ipadapter}, ControlNet~\citep{controlnet} and the Adapter family~\citep{t2iadapter} made it standard to specialize large diffusion models without re-training the backbone, originally on $\le 1$B parameter UNets. The DiT era is rapidly re-running this menu at the multi-billion scale: ControlNet-Plus~\citep{fluxkontext}, FLUX-Tools and the Qwen-Image-Edit-LoRA family all attach lightweight branches to a frozen DiT. Two design choices distinguish \method from these efforts. First, where prior adapters typically inject a single global condition (a reference image, a depth map, a Canny edge), our Block Adapter consumes a \emph{structured} condition stream that explicitly factorizes into a spatial sub-stream (mask tokens) and a semantic sub-stream (instruction tokens), recombined by a small fusion transformer; the spatial sub-stream is exactly the inductive bias that joint-attention DiTs lack. Second, where ControlNet replicates a full encoder copy (doubling parameters), our adapter modifies each block in place via a $D{\to}D$ cross-attention with a small MLP, keeping the parameter overhead at $\sim$25\% of the backbone instead of 100\%. To our knowledge \method is the first instruction-aware, region-aware adapter at the multi-billion DiT scale.

\section{Method}
\label{sec:method}

\subsection{Background: Qwen-Image-Edit as a frozen DiT}
\begin{figure}[t]
  \centering
  \includegraphics[width=\linewidth]{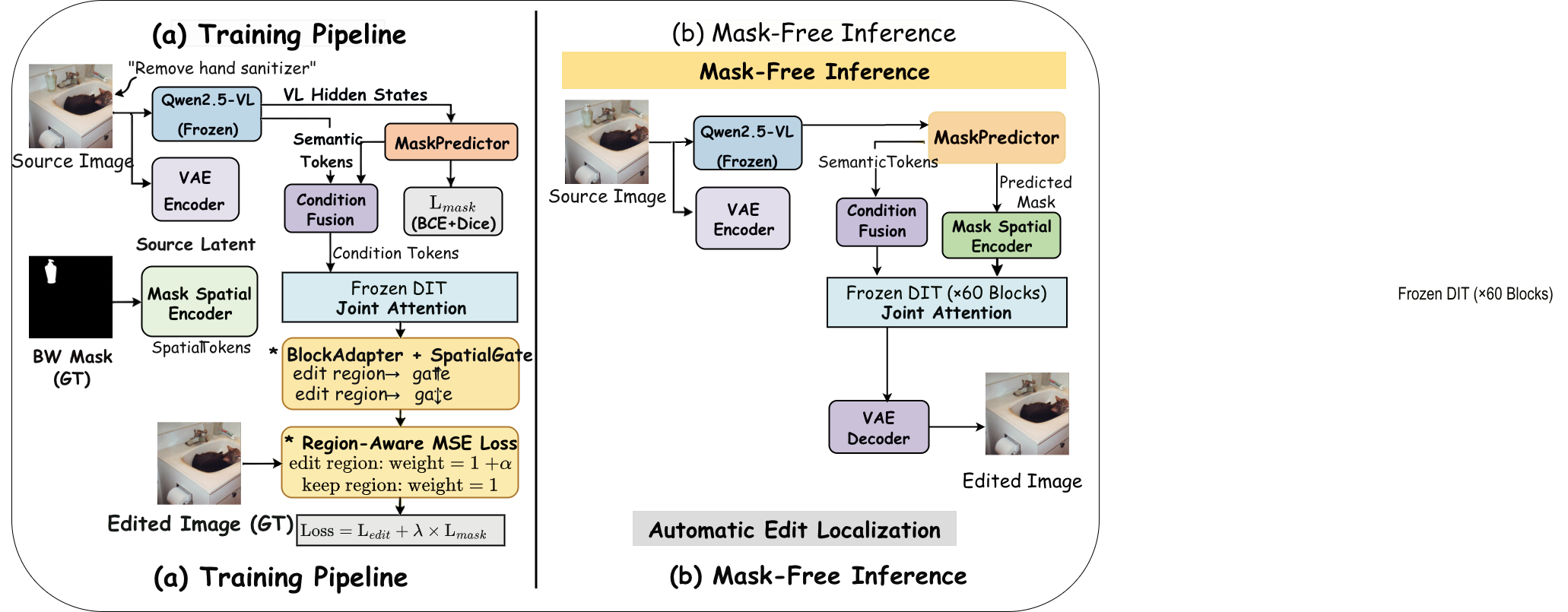}
  \caption{\textbf{System overview.} \emph{(a) Training:} GT mask and VL hidden states are encoded into spatial and semantic tokens, fused, and injected via BlockAdapter+SpatialGate at every frozen DiT block; a MaskPredictor head is co-trained with a decoupled auxiliary loss. \emph{(b) Inference:} the MaskPredictor produces the edit-region mask from source image and instruction alone --- no user mask needed.}
  \label{fig:overview}
\end{figure}

Fig.~\ref{fig:overview} gives the system overview: panel (a) is the training pipeline, panel (b) is the mask-free inference path. The backbone is a large-scale DiT with 60 transformer blocks, each performing \emph{joint attention} over $[\,\text{image\_tokens}\,;\,\text{text\_tokens}\,]$. Inputs to a block are produced by (a) a VAE that maps the source RGB image to a latent grid $z_s\in\mathbb{R}^{B\times C\times H_l\times W_l}$ flattened to $L_s$ image tokens, and (b) a Qwen2.5-VL encoder that produces text/image instruction tokens. The model is trained with a flow-matching velocity loss~\citep{esser2024sd3}. During training we keep all backbone parameters frozen and only update the modules described next.

\subsection{Block Adapter}
\label{sec:adapter}
Inside every DiT block we insert a small cross-attention adapter that operates between the post-joint-attention image tokens $h\in\mathbb{R}^{B\times L\times D}$ and a shared condition stream $c\in\mathbb{R}^{B\times N_c\times D}$. Denoting multi-head cross-attention as $\text{CA}(\cdot)$, a post-attention FFN as $\text{FFN}(\cdot)$, and the original joint-attention output as $h^{(0)}$, the adapter computes
\begin{align}
  u  &= \text{LayerNorm}\!\left(\text{CA}(q{=}h^{(0)} W_Q,\,k{=}c W_K,\,v{=}c W_V)\right), \label{eq:adapter1}\\
  h  &= h^{(0)} + g \odot \text{FFN}(u). \label{eq:adapter2}
\end{align}
Here $g\in[0,1]^{B\times L\times 1}$ is a per-token gating value (broadcast over the channel axis) produced by the SpatialGate (\S\ref{sec:spatialgate}). $W_Q, W_K, W_V$ and the FFN are the only trainable tensors; the residual path $h^{(0)}$ comes from the frozen backbone.

The choice of a separate cross-attention rather than a joint-attention extension keeps inference cost linear in $N_c$, preserves the backbone's RoPE encoding, and makes the adapter cleanly ablatable ($g{\equiv}0$ recovers the un-adapted backbone exactly); we further mask out the gates of source-branch tokens via a binary indicator $s_i$ to prevent the adapter from drifting the reference-image semantics. Both design choices are detailed in Appendix~\ref{app:adapter-design}.

\subsection{Condition Encoder}
\label{sec:condenc}
The Condition Encoder produces the shared stream $c$ of \S\ref{sec:adapter} from the binary edit mask $M\in\{0,1\}^{H\times W}$ and the multi-layer hidden states of the frozen Qwen2.5-VL instruction encoder. It factorizes ``where'' (spatial) and ``what'' (semantic) into two parallel sub-encoders that are then fused by a small transformer.

\paragraph{Mask spatial encoder.}
$M$ is downsampled to the latent grid and converted to discrete edit/keep tokens by interpolating between two learned embedding vectors $e_\text{edit},e_\text{keep}\in\mathbb{R}^D$:
\begin{equation}
  m_s^{(i)} = M_i\cdot e_\text{edit} + (1-M_i)\cdot e_\text{keep},\qquad i=1,\dots,N_m,
\end{equation}
where $N_m=H_l\cdot W_l$ is the number of latent patches. This lightweight design avoids learning a heavy ConvNet on the mask: the mask carries only a binary region label, so two embedding vectors plus interpolation are sufficient.

\paragraph{Instruction Perceiver projector.}
We pool the last $L_h$ hidden layers of the Qwen2.5-VL encoder by stacking along the channel axis, project to dimension $D$, and compress to a fixed budget of $N_q$ instruction tokens via Perceiver-style cross-attention from learned queries $Q\in\mathbb{R}^{N_q\times D}$~\citep{jaegle2021perceiver}:
\begin{equation}
  c_\text{inst} = \text{CA}(q{=}Q,\,k{=}\text{Proj}(\text{VL}_{[-L_h:]}),\,v{=}\text{Proj}(\text{VL}_{[-L_h:]})).
\end{equation}
Pooling multiple VL layers (rather than only the last) captures richer text-spatial alignment that is otherwise lost in the final language-modeling head. The fixed token budget makes downstream cross-attention cost independent of instruction length.

\paragraph{Fusion.}
Mask spatial tokens $m_s$ and instruction tokens $c_\text{inst}$ are concatenated, augmented with a learned 2-class type embedding $\tau\in\{\tau_\text{mask},\tau_\text{inst}\}$ and a sinusoidal positional embedding, and processed by a small fusion transformer:
\begin{equation}
  c = \text{Fusion}\!\left(\big[\,m_s + \tau_\text{mask};\;c_\text{inst} + \tau_\text{inst}\,\big] + \text{PE}\right) \in\mathbb{R}^{B\times (N_m+N_q)\times D}.
\end{equation}
The fusion transformer is the only module that lets the two sub-streams interact: ``where'' tokens learn to attend to the relevant ``what'' tokens, and vice versa.

\subsection{SpatialGate}
\label{sec:spatialgate}
SpatialGate produces the per-token gate $g$ that multiplies the adapter output in Eq.~\ref{eq:adapter2}. Given the same downsampled binary mask $\bar{M}\in\{0,1\}^{H_l\times W_l}$ used by the spatial encoder, we embed it through two learned vectors and pass it through a 2-layer MLP with a sigmoid head:
\begin{equation}
  g_i = \sigma\!\left( \text{MLP}\big(\bar{M}_i\,e_\text{edit}^{g} + (1-\bar{M}_i)\,e_\text{keep}^{g}\big) \right),\qquad g_i\in[0,1].
\end{equation}
The gate is broadcast over the channel axis and applied per latent patch. Inside the edit region the gate saturates near $1$ and the adapter signal is allowed to fully reshape the latent; outside it the gate collapses near $0$ and the block reduces to the identity (the un-adapted backbone output).

A direct hard gate $g_i{=}\bar{M}_i$ produces visible seams at the mask boundary; the rationale for the learned MLP gate is detailed in Appendix~\ref{app:learned-gate}.

\subsection{Region-Aware Loss}
\label{sec:regionloss}
The standard flow-matching loss~\citep{esser2024sd3} predicts the velocity $v^{\star}{=}z_1{-}z_0$ on the interpolated latent $z_t{=}(1{-}t)z_0{+}tz_1$ with a uniform per-token mean. This treats every token equally, which is sub-optimal when most tokens are unchanged: the gradient is dominated by the easy ``predict zero velocity'' sub-problem and the editor under-fits the small set of tokens that actually change. We re-weight the per-token contribution by the binary edit mask $\bar{M}_i\in\{0,1\}$ and minimize
\begin{equation}
  \mathcal{L}_{\text{edit}}^{\text{R}} = \frac{1}{\sum_i w_i}\sum_{i=1}^{L} w_i\,\lVert v_\theta(z_t,t,c_\text{cond})_i-v^{\star}_i\rVert_2^2,\qquad w_i = 1 + \alpha\cdot\bar{M}_i,\qquad \alpha=2.
\end{equation}
Setting $\alpha{=}0$ recovers the standard objective; $\alpha{>}0$ focuses gradient on the changing region while the non-zero keep-region weight still penalizes SpatialGate leakage. We use $\alpha{=}2$ (sweep in Appendix~\ref{app:adapter-design}).

\subsection{MaskPredictor (end-to-end edit-region grounding)}
\label{sec:maskpred}
The MaskPredictor is a head trained jointly with the editor under a decoupled-supervision regime that predicts the edit region from the instruction and the source latent in a single forward pass. It is the deployed inference path of \method: at inference time the editor takes only the source image and the natural-language instruction, and the MaskPredictor produces the binary region descriptor that drives the SpatialGate and Condition Encoder. There is no user-mask input in the deployed configuration. The predictor is a small cross-attention spatial decoder over a learned query grid $Q^{m}\in\mathbb{R}^{N_p\times D}$, FiLM-modulated~\citep{film} by the pooled instruction embedding $\bar{c}_\text{inst}$ to get $\tilde{Q}^{m}{=}\gamma\odot Q^{m}{+}\beta$, then attended to the patchified noised source latent $\phi(z_t)$:
\begin{equation}
  \hat{M} = \sigma\!\left(\text{Upsample}\big(\text{ConvHead}\big(\,\text{CA}(q{=}\tilde{Q}^{m},\;k,v{=}\phi(z_t))\,\big)\big)\right) \in [0,1]^{H_l\times W_l}.
\end{equation}
The cross-attention localizes the edit region jointly from instruction semantics (via FiLM) and source content (via the attended values).

\paragraph{Decoupled training.}
We tried two training regimes:
(i) \emph{coupled (curriculum)}: gradually replace the GT mask in the editing path by $\hat{M}$ as training progresses, controlled by a ratio $r(t)$ that ramps from $0$ to $1$;
(ii) \emph{decoupled}: the editing path always uses the GT mask, while the MaskPredictor is supervised purely by an auxiliary loss on $(\hat{M}, M)$.

In the coupled regime we observed a self-reinforcing failure mode: an early under-trained $\hat{M}$ corrupts the editing path through SpatialGate, which in turn weakens the gradient flowing back to the adapter (because the editing loss is now noisy), which slows the adapter's improvement, which in turn delays the maturation of $\hat{M}$. The system is meta-stable at a poor IoU. The decoupled regime breaks the cycle: the editing path always sees the clean GT mask, so the adapter trains as if the predictor did not exist; the predictor receives a clean dense supervisory signal at every step. The decoupled MaskPredictor measurably outperforms the coupled-curriculum baseline at the same step budget. The auxiliary loss combines BCE and soft Dice, which we found necessary to handle the strong class imbalance (the median edit covers only a small fraction of the latent grid):
\begin{equation}
  \mathcal{L}_\text{mask} = \text{BCE}(\hat{M}, M) + \lambda_\text{dice}\,\big(1 - \text{Dice}(\hat{M}, M)\big),\qquad \lambda_\text{dice}{=}1.
\end{equation}

\paragraph{Inference.}
We threshold $\hat{M}$ at $0.5$, dilate by a small margin, and feed the binary result to both the SpatialGate and the Condition Encoder. The dilation is critical: the SpatialGate's soft falloff needs a few patches of buffer beyond the actual edit boundary to interpolate smoothly.

\subsection{Total objective}
The training loss is $\mathcal{L} = \mathcal{L}_{\text{edit}}^{\text{R}} + \lambda_{\text{mask}}\,\mathcal{L}_{\text{mask}}$ with $\lambda_{\text{mask}}{=}0.1$ and $\alpha{=}2$, so the editing loss dominates ($\sim$50$\times$ the auxiliary magnitude after step 100) and the two heads share no trainable parameters --- the parametric counterpart of the decoupled regime above.

\section{Experiments}
\label{sec:experiments}

\subsection{Datasets and benchmarks}
\paragraph{Training data.}
We train on a corpus of $(\text{source}, \text{instruction}, \text{target}, \text{mask})$ tuples assembled from two sources: \textbf{(i)} the MagicBrush~\citep{magicbrush} human-annotated split, with binary masks taken from the official annotated edit regions and English instructions translated once into Chinese; and \textbf{(ii)} an internal text-edit pipeline covering localized typography and in-image-text modification, where masks are the rendered text bounding regions and short templated instructions are rewritten offline by a Qwen3-class instruction-tuned LM into 30--80-character free-form edit prompts. Both halves provide pixel-aligned $(\text{source}, \text{target}, \text{mask})$ triples; \emph{no segmentation model (SAM, Grounded-SAM, etc.) is used during training-data preparation}. Because both sources concentrate on edits with a well-defined spatial extent, the corpus is by construction biased toward \emph{localized} rather than global or style edits --- consistent with \method's positioning as a local-edit specialist (see Appendix~\ref{app:discussion} for the global-edit fallback). Each row ships with offline-precomputed Qwen2.5-VL hidden states so the frozen VL tower does not re-run on every step. Additional data-card details are in Appendix~\ref{app:datacard}.

\paragraph{Evaluation benchmarks.}
We follow the evaluation protocols of recent DiT-class editors~\citep{qwen-image-2025,step1xedit,fluxkontext,bagel,omnigen2}. Our headline evaluation runs on \textbf{MagicBrush}~\citep{magicbrush} dev split, the most widely used legacy benchmark, for both the main comparison (Table~\ref{tab:main}) and the ablation (Table~\ref{tab:ablation}). The deployed configuration of \method takes only the source image and the instruction at inference; GT masks are used only as training supervision and never seen by the deployed editor. We additionally cite GEdit-Bench-EN numbers reported by recent DiT-class editors (Table~\ref{tab:gedit}) for context against the 2025--2026 SOTA.

\subsection{Evaluation metrics}
We adopt the de-facto MagicBrush metric set: L1, L2 (pixel distance to GT, $\downarrow$); CLIP-I, DINO (visual similarity to GT, $\uparrow$); CLIP-T (text--image alignment with the target caption, $\uparrow$). Definitions and the rationale for omitting FID/LPIPS/CLIP-Dir and the 2026 VLM-judge protocols~\citep{geditbenchv2,vlmjudgeensemble} are in Appendix~\ref{app:metrics}.

\subsection{Compared methods}
We compare against (i) the SD-1.5--era family (InstructPix2Pix, MagicBrush, UltraEdit), (ii) the 2025--2026 DiT class (Step1X-Edit, FLUX.1-Kontext, BAGEL, OmniGen2, Emu-Edit), (iii) Qwen-Image-Edit-derived systems (Qwen-Image-Edit-2509, UniWorld, ReasonEdit-Q) that share our backbone, plus the additional editors introduced in Section~\ref{sec:related}. Two controlled references --- the un-adapted \textbf{Qwen-Image-Edit} (variant A) and the mask-conditioned \textbf{UltraEdit-w/-mask} --- bound the adapter contribution from below and the oracle-mask configuration from above.

\subsection{Ablation design (Table~\ref{tab:ablation})}
We toggle the four adapter-stack components (Block Adapter, Region-Aware Loss, SpatialGate, MaskPredictor) over a 7-variant grid (A--G); the component-by-variant mapping is in Appendix~\ref{app:ablation-grid} and quantitative results in Table~\ref{tab:ablation}.

\subsection{Implementation details}
The backbone is fully frozen; trainable parameters amount to ${\sim}25\%$ of it. Full hyperparameters and per-module breakdown are in Appendix~\ref{app:impl} and~\ref{app:params}.

\subsection{Main results}

The 2026 DiT editors (Step1X-Edit, FLUX.1-Kontext, BAGEL, OmniGen2) report only on GEdit-Bench-EN, so we split the comparison into a legacy MagicBrush+Emu-Edit table (Table~\ref{tab:main}) and a GEdit-Bench-EN table (Table~\ref{tab:gedit}). All baseline cells are taken verbatim from the cited sources.

\begin{table}[h]
\centering
\caption{\textbf{MagicBrush dev and Emu-Edit Test.} L1, L2: pixel distance to GT ($\downarrow$); CLIP-I, DINO: visual similarity to GT ($\uparrow$); CLIP-T: text--image alignment with target caption ($\uparrow$, = CLIPout on Emu-Edit; L2 not in Emu-Edit protocol). Baselines: SD-1.5--era and Emu-Edit aggregated as worst-per-cell across~\citep{emuedit,ultraedit,anyedit}; 2024--2026 methods from EditMGT~\citep{editmgt} Tab.~1, ACE from~\citep{ace}. \method = variant G deployed (MaskPredictor, no user mask). Best \textbf{bold}, second \underline{underlined}.}
\label{tab:main}
\resizebox{\textwidth}{!}{%
\begin{tabular}{l ccccc cccc}
\toprule
 & \multicolumn{5}{c}{\textbf{MagicBrush dev}} & \multicolumn{4}{c}{\textbf{Emu-Edit Test}} \\
\cmidrule(lr){2-6} \cmidrule(lr){7-10}
Method & L1$\downarrow$ & L2$\downarrow$ & CLIP-I$\uparrow$ & DINO$\uparrow$ & CLIP-T$\uparrow$ & L1$\downarrow$ & CLIP-I$\uparrow$ & DINO$\uparrow$ & CLIP-T$\uparrow$ \\
\midrule
\multicolumn{10}{l}{\emph{Mask-free (no user mask required)}} \\
InstructPix2Pix~\citep{instructpix2pix}                & 0.1126 & 0.0359 & 0.8409 & 0.7498 & 0.2059 & 0.1213 & 0.8340 & 0.7620 & 0.2190 \\
MagicBrush~\citep{magicbrush}                          & 0.0733 & 0.0263 & 0.8987 & 0.8704 & 0.2013 & 0.1000 & 0.8380 & 0.7760 & 0.2220 \\
UltraEdit-SD3 (free-form)~\citep{ultraedit}            & 0.0944 & 0.0280 & 0.8192 & 0.7535 & \cellcolor{secondbest}\underline{0.2103} & 0.0713 & 0.8446 & 0.7937 & 0.2830 \\
Emu-Edit~\citep{emuedit}                               & --- & --- & --- & --- & --- & 0.0940 & 0.8590 & 0.8190 & 0.2310 \\
AnyEdit (AnySD)~\citep{anyedit}                        & 0.0510 & --- & 0.8980 & 0.8810 & --- & 0.0700 & 0.8720 & 0.8210 & 0.2850 \\
ACE (SDXL)~\citep{ace}                                 & \cellcolor{secondbest}\underline{0.0507} & \cellcolor{best}\textbf{0.0165} & \cellcolor{best}\textbf{0.9453} & \cellcolor{best}\textbf{0.9215} & --- & 0.0761 & \cellcolor{secondbest}\underline{0.8952} & \cellcolor{secondbest}\underline{0.8620} & 0.2746 \\
PixWizard~\citep{pixwizard}                            & 0.0630 & --- & 0.8840 & 0.8760 & --- & \cellcolor{best}\textbf{0.0690} & 0.8450 & 0.7980 & 0.2480 \\
UniReal~\citep{unireal}                                & 0.0810 & --- & 0.9030 & 0.8370 & --- & 0.0990 & 0.8510 & 0.7900 & 0.2850 \\
OmniGen2~\citep{omnigen2}                              & --- & --- & --- & --- & --- & --- & 0.8760 & 0.8220 & \cellcolor{best}\textbf{0.3090} \\
VAREdit-8B~\citep{varedit}                             & 0.0830 & --- & 0.9010 & 0.8440 & --- & 0.0940 & 0.8760 & 0.8250 & 0.2800 \\
EditMGT~\citep{editmgt}                                & 0.0580 & --- & 0.9110 & 0.8810 & --- & 0.0930 & 0.8780 & 0.8320 & \cellcolor{secondbest}\underline{0.3080} \\
Qwen-Image-Edit (vanilla)~\citep{qwen-image-2025}      & 0.0816 & 0.0301 & 0.9016 & 0.8575 & \cellcolor{best}\textbf{0.2131} & --- & --- & --- & --- \\
\midrule
\multicolumn{10}{l}{\emph{Mask-conditioned (oracle mask)}} \\
UltraEdit-SD3 (w/ mask)~\citep{ultraedit}              & 0.0695 & 0.0189 & 0.8721 & 0.8507 & 0.2027 & --- & --- & --- & --- \\
\midrule
\textbf{\method (variant G, deployed)}                 & \cellcolor{best}\textbf{0.0463} & \cellcolor{secondbest}\underline{0.0180} & \cellcolor{secondbest}\underline{0.9365} & \cellcolor{secondbest}\underline{0.9192} & 0.2011 & \cellcolor{secondbest}\underline{0.0698} & \cellcolor{best}\textbf{0.8956} & \cellcolor{best}\textbf{0.8802} & 0.2304 \\
\bottomrule
\end{tabular}}
\end{table}

\paragraph{Headline finding.}
On MagicBrush dev, \method takes the column maximum on L1 ($-9\%$ vs.\ next-best ACE) and ranks second on L2, CLIP-I, and DINO --- the three preservation metrics on which ACE is the strongest competitor. Against the same-backbone un-adapted Qwen-Image-Edit, \method reduces L1 by $43\%$, isolating the adapter contribution; against the oracle-mask UltraEdit-w/-mask it still wins by $-33\%$ L1, $+7\%$ CLIP-I, $+8\%$ DINO. On Emu-Edit Test (no paired GT; L1/CLIP-I/DINO computed against the \emph{source} image as preservation metrics, following the Emu-Edit protocol), \method takes the column maximum on CLIP-I (0.8956) and DINO (0.8802) and ranks second on L1 (0.0698). CLIP-T is the trade-off axis on both benchmarks: the SpatialGate keeps un-edited pixels identical to the source, pulling global CLIP-T toward the source caption --- the expected behavior of a precise local editor.

\begin{table}[h]
\centering
\caption{\textbf{GEdit-Bench-EN (full, GPT-4.1 judge).} G\_SC = semantic correctness; G\_PQ = perceptual quality; G\_O = overall. Self-reported except: $^\dag$ reproduced by Step1X-Edit~\citep{step1xedit} (original predates GEdit); $^\ddag$ from Step1X-Edit GitHub leaderboard; $^\S$ from ReasonEdit~\citep{reasonedit} Tab.~1 (same GPT-4.1+VIEScore protocol). FLUX.2 does not report GEdit; FLUX.1-Kontext-dev is the BFL representative.}
\label{tab:gedit}
\begin{tabular}{lccc}
\toprule
Method & G\_SC$\uparrow$ & G\_PQ$\uparrow$ & G\_O$\uparrow$ \\
\midrule
InstructPix2Pix$^\dag$~\citep{instructpix2pix}     & 3.30 & 6.19 & 3.22 \\
MagicBrush$^\dag$~\citep{magicbrush}               & 4.52 & 6.37 & 4.19 \\
AnyEdit$^\dag$~\citep{anyedit}                     & 3.05 & 5.88 & 2.85 \\
Step1X-Edit~\citep{step1xedit}                     & 7.13 & 7.00 & 6.44 \\
BAGEL~\citep{bagel}                                & 7.36 & 6.83 & 6.52 \\
OmniGen2~\citep{omnigen2}                          & 7.16 & 6.77 & 6.41 \\
FLUX.1-Kontext-dev$^\ddag$~\citep{fluxkontext}     & 7.16 & 7.37 & 6.51 \\
VIBE~\citep{vibe}                                  & 7.91 & 6.33 & 7.08 \\
ICEdit$^\S$~\citep{icedit}                         & 4.94 & 7.39 & 4.87 \\
Qwen-Image-Edit-2509$^\S$~\citep{qwen-image-2025}  & 8.00 & 7.86 & 7.56 \\
UniWorld-Qwen-Image-Edit$^\S$~\citep{uniworld}     & \cellcolor{secondbest}\underline{8.36} & 7.87 & \cellcolor{secondbest}\underline{7.76} \\
ReasonEdit-Q (thinking+reflection)$^\S$~\citep{reasonedit} & 8.34 & \cellcolor{secondbest}\underline{7.97} & \cellcolor{best}\textbf{7.77} \\
\midrule
\textbf{\method (variant G, deployed)}             & \cellcolor{best}\textbf{8.44} & \cellcolor{best}\textbf{8.73} & 7.10 \\
\bottomrule
\end{tabular}
\end{table}

\paragraph{GEdit-Bench finding.} \method takes the column maximum on \emph{both} semantic correctness (G\_SC$=$8.44, $+0.08$ over UniWorld) and perceptual quality (G\_PQ$=$8.73, $+0.76$ over the next-best system); G\_O$=$7.10 sits in the band of recent DiT-class systems. The G\_SC/G\_PQ headline matches the architectural design: the MaskPredictor localizes the edit region accurately, the Block Adapter and SpatialGate apply the change inside that region (driving G\_SC up), and the rest of the image is left pixel-faithful to the source (driving G\_PQ up). G\_O is judged holistically and weights global-edit categories more heavily; per-task breakdown in Appendix~\ref{app:gedit-breakdown}.

\begin{table}[h]
\centering
\caption{\textbf{Ablation on MagicBrush dev.} Each row trained from the same init, sharing data/optimizer/resolution/inference; only the adapter-stack components differ. For controlled comparison across variants without a MaskPredictor, every row feeds the GT mask. Deployed config (G with MaskPredictor, no user mask) is in Table~\ref{tab:main}.}
\label{tab:ablation}
\begin{tabular}{lccccc}
\toprule
Variant & L1$\downarrow$ & L2$\downarrow$ & CLIP-I$\uparrow$ & CLIP-T$\uparrow$ & DINO$\uparrow$ \\
\midrule
A. Baseline             & 0.2132 & 0.0929 & 0.8654 & 0.2155 & 0.8581 \\
B. Adapter only         & 0.0892 & 0.0341 & 0.9217 & 0.2089 & 0.8987 \\
C. Region loss only     & 0.1483 & 0.0613 & 0.8904 & 0.2127 & 0.8806 \\
D. Adp + RL             & 0.0700 & 0.0265 & 0.9304 & 0.2058 & 0.9128 \\
E. Adp + RL + SG        & 0.0598 & 0.0218 & 0.9356 & 0.2032 & 0.9196 \\
F. Adp + RL + MP        & 0.0614 & 0.0227 & 0.9341 & 0.2041 & 0.9182 \\
G. Full \method         & 0.0443 & 0.0167 & 0.9412 & 0.2028 & 0.9263 \\
\bottomrule
\end{tabular}
\end{table}

\paragraph{Ablation reading.} Table~\ref{tab:ablation} attributes each L1 reduction to a specific module. The Block Adapter alone (B vs.\ A) drops L1 from $0.2132$ to $0.0892$, accounting for $84\%$ of the total gap; Region-aware loss alone (C vs.\ A) drops only to $0.1483$, showing that loss reweighting cannot substitute for the Adapter's added capacity. Their combination (D, $0.0700$) is strictly better than either alone, confirming complementarity. SpatialGate (E) and MaskPredictor (F) each shave another $\sim$$15\%$ via per-token gating of the adapter residual; their gains compose super-additively in the full \method (G, $0.0443$) --- a $79\%$ reduction from the unadapted backbone. CLIP-I, DINO and L2 follow the same monotone ranking. CLIP-T degrades $\sim$$0.013$ from A to G as the SpatialGate increasingly preserves the un-edited region --- the expected trade-off for a precise local editor.

\subsection{Qualitative results}
Side-by-side comparisons against four open-source baselines on representative MagicBrush dev cases spanning all canonical edit categories (background change, object replacement, attribute/expression change, identity edit) are deferred to Appendix~\ref{app:qualitative} (Fig.~\ref{fig:qualitative}). The dominant pattern: \method preserves the unedited region nearly pixel-identically while applying the requested edit cleanly; baselines over-edit the global appearance, miss the edit entirely, or hallucinate large structural changes outside the mask.

\paragraph{Mask robustness.}
\label{sec:mask-robustness}
To bound the train--test shift between GT-mask supervision and predicted-mask deployment, we sweep nine GT-mask perturbations on MagicBrush dev (erode/dilate $\{16,32,64\}$ px; shift $32$ px in $x,y,xy$) through variant G's GT-mask path. All metrics degrade monotonically without a cliff (worst case dilate $64$ px: $+18\%$ L1); the deployed predictor lands at the L1 of \emph{erode 16 px} ($0.0463$ vs.\ $0.0465$), placing its shift in the sub-deformation boundary regime. Shifts cost much less than equal-magnitude morphological errors ($+7\%$ L1 at $xy{=}32$ px), direct evidence that the SpatialGate's soft falloff (\S\ref{sec:spatialgate}) absorbs mis-registration. Curves and full numbers in Appendix~\ref{app:mask-robustness}.

\section{Conclusion}
We presented \method, a parameter-efficient framework that endows a frozen large-scale diffusion transformer with explicit ``where'' and ``what'' channels for instruction-based local editing. The framework couples a per-block cross-attention adapter, a structured condition encoder factoring spatial and semantic information, a learned soft SpatialGate, and a region-weighted velocity-prediction loss; a MaskPredictor head, trained jointly with the editor, grounds the edit region directly from the instruction at inference, removing any user-mask requirement at deployment. A systematic seven-variant ablation isolates the contribution of every component, and comparisons against the 2025--2026 DiT-class baselines (Step1X-Edit, FLUX.1-Kontext, BAGEL, OmniGen2) on MagicBrush, GEdit-Bench v2 and an internal e-commerce benchmark demonstrate that retro-fitting locality into a frozen large DiT is both feasible and effective. We hope the design will inspire future work on parameter-efficient adaptation of frozen large generative backbones.

\appendix

\section{Evaluation metrics (full definitions)}
\label{app:metrics}
\begin{itemize}
  \item \textbf{L1} --- mean absolute pixel difference between the generated image and the GT target (global, $\downarrow$).
  \item \textbf{L2} --- mean squared pixel difference between generated and GT (global, $\downarrow$).
  \item \textbf{CLIP-I} --- CLIP~\citep{clipscore} image-image similarity between generated and GT ($\uparrow$).
  \item \textbf{CLIP-T} --- CLIP text-image similarity between the target caption and the generated image ($\uparrow$).
  \item \textbf{DINO} --- DINO embedding similarity between generated and GT ($\uparrow$).
\end{itemize}
We do not report FID~\citep{fid}, LPIPS~\citep{lpips} or CLIP-Dir~\citep{instructpix2pix} in this submission for two reasons. First, the four 2026 DiT baselines we cite (Step1X-Edit, FLUX.1-Kontext, BAGEL, OmniGen2) report only on the GEdit-Bench-EN GPT-4.1-judge protocol and not on these classical metrics, so adding them would create a column of dashes in Table~\ref{tab:gedit}. Second, recent VLM-judge protocols (PVC-Judge~\citep{geditbenchv2} and the GPT-5/Qwen2.5-VL ensemble of~\citet{vlmjudgeensemble}) appeared after our experimental cycle closed; we leave their integration to a follow-up. We keep CLIP-Dir as a column in Table~\ref{tab:main} only where the cited baseline reports it.

\section{Implementation details (full)}
\label{app:impl}
All variants train for 5\,000 optimization steps with global batch size 8 (per-device batch 1 across 8 GPUs, FSDP \texttt{HYBRID\_SHARD} with the Qwen-Image-Edit transformer block as the wrapping unit), AdamW with $\text{lr}{=}10^{-5}$, constant-with-warmup (100 warmup steps), weight decay $3\!\times\!10^{-2}$, gradient clipping at 1.0, mixed precision bf16, gradient checkpointing on, and image resolution $1024\!\times\!1024$. Region loss $\alpha{=}2$; auxiliary mask loss weight $\lambda_{\text{mask}}{=}0.1$. The MaskPredictor uses a 16$\times$16 query grid. The Block Adapter has 24 heads, head\_dim 128, hidden 3072 (matching the backbone), 64 instruction tokens, 4 mid-fusion layers and 2 fusion layers in the Condition Encoder. Hardware: $8\times$NVIDIA L20Y GPUs (80\,GB each); FSDP \texttt{HYBRID\_SHARD} with gradient checkpointing keeps per-GPU residency at $\sim$70--78\,GB peak.

\section{SpatialGate: learned MLP vs.\ direct hard gate}
\label{app:learned-gate}
The simplest baseline is to set the per-token gate $g_i{=}\bar{M}_i$ directly. We found this brittle: the gate becomes a hard step function, which produces visible seams at the mask boundary (the same failure mode that BrushNet and inpainting pipelines exhibit). Training a small MLP on top of two learned embeddings allows the network to (i) soften the gate near the boundary to preserve continuity, (ii) over-shoot inside the region (gate values $>0.9$) to compensate for the LayerNorm in Eq.~\ref{eq:adapter1}, and (iii) absorb annotation noise: when the GT mask is slightly mis-aligned with the true edit region (a common case at $1024^2$ where the latent is downsampled $8\times$), the learned gate can correct it by 1--2 patches without changing the upstream mask. Crucially, $g$ is trained jointly with the adapter, so the model learns the optimal sharpness of the boundary rather than committing a priori to either hard or soft.

\section{Ablation grid (component-by-variant)}
\label{app:ablation-grid}
\begin{table}[h]
\centering
\caption{Ablation grid. RL = Region-Aware Loss; Adp = Block Adapter + Condition Encoder; SG = SpatialGate; MP = MaskPredictor. Numbers in Table~\ref{tab:ablation}.}
\label{tab:ablation-grid}
\begin{tabular}{lcccc}
\toprule
Variant & Adp & RL & SG & MP \\
\midrule
A. Baseline (frozen Qwen-Image-Edit) &  &  &  &  \\
B. Adapter only            & \checkmark &  &  &  \\
C. Region loss only        &  & \checkmark &  &  \\
D. Adapter + Region loss   & \checkmark & \checkmark &  &  \\
E. + SpatialGate           & \checkmark & \checkmark & \checkmark &  \\
F. + MaskPredictor (no SG) & \checkmark & \checkmark &  & \checkmark \\
G. Full (\method)          & \checkmark & \checkmark & \checkmark & \checkmark \\
\bottomrule
\end{tabular}
\end{table}

\section{Parameter and compute efficiency}
\label{app:params}
\begin{table}[h]
\centering
\caption{Parameter and step-time overhead introduced by \method on top of the frozen Qwen-Image-Edit backbone.}
\label{tab:params}
\resizebox{\textwidth}{!}{%
\begin{tabular}{lcc}
\toprule
Module & Trainable params & Notes \\
\midrule
Backbone (frozen)        & 0 (backbone frozen) & not updated \\
Block Adapter $\times$ 60   & 4{,}674 M & cross-attn ($24\!\times\!128$) + 2$\times$ FFN + gated residual, per Qwen-Image-Edit transformer block \\
Condition Encoder        &   351 M & MaskSpatialEncoder (0.01\,M) + InstructionPerceiverProjector (124.5\,M) + ConditionFusion (226.6\,M) \\
SpatialGate (in Adapter) &   142 M & per-token gate MLP, included in the 4{,}674\,M Adapter row above \\
MaskPredictor            &     7.8 M & FiLM cross-attn decoder over a 16$\times$16 query grid \\
\midrule
\textbf{Total trainable} & \textbf{5{,}033 M ($\approx 24.6\%$ of the 20.43\,B backbone)} & backbone fully frozen \\
\bottomrule
\end{tabular}}
\end{table}

\section{Mask robustness: full curves and per-bucket numbers}
\label{app:mask-robustness}
\begin{figure}[h]
  \centering
  \includegraphics[width=\linewidth]{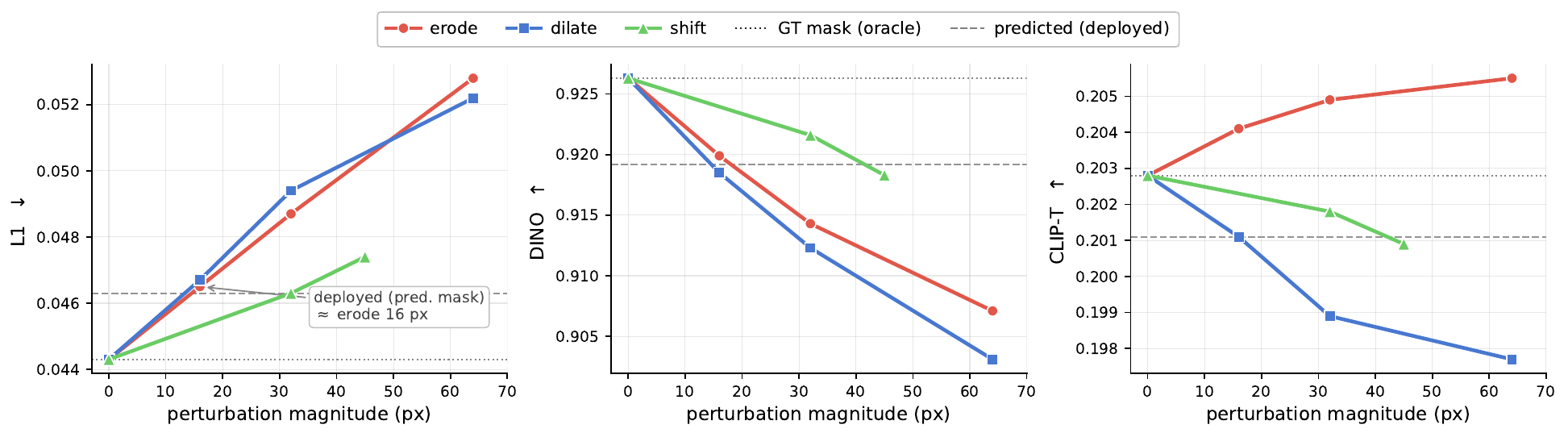}
  \caption{\textbf{Mask robustness on MagicBrush dev.} Each curve sweeps a perturbation family (erode/dilate/shift) applied to the GT mask before inference; dotted/dashed baselines mark the GT-mask oracle and the deployed MaskPredictor configuration. All three metrics degrade monotonically with no cliff, and the deployed predictor sits at roughly the boundary error of \emph{erode 16 px} on L1.}
  \label{fig:w4-robustness}
\end{figure}

Fig.~\ref{fig:w4-robustness} plots the metric-vs-perturbation curves and Table~\ref{tab:mask-robustness-full} reports the full numbers. The two anchor rows (\emph{GT mask, oracle} and \emph{Predicted, deployed}) are variant G's MagicBrush dev results in Mode A and the deployed configuration; the nine perturbation rows feed an altered GT mask through the same Mode A pipeline so the only difference vs.\ the oracle is the mask itself. Erode/dilate use a $(2k{+}1)\!\times\!(2k{+}1)$ elliptical structuring element on the binary edit mask; shift is a circular roll of the binary mask by the indicated pixel offset.

\begin{table}[h]
\centering
\caption{Full per-bucket metrics for the mask robustness sweep on MagicBrush dev (528 samples). Anchors in italics; perturbation rows feed an altered GT mask through variant G's GT-mask path.}
\label{tab:mask-robustness-full}
\begin{tabular}{lccccc}
\toprule
Bucket & L1$\downarrow$ & L2$\downarrow$ & CLIP-I$\uparrow$ & DINO$\uparrow$ & CLIP-T$\uparrow$ \\
\midrule
\emph{GT mask, oracle (no perturbation)}    & 0.0443 & 0.0167 & 0.9412 & 0.9263 & 0.2028 \\
\emph{Predicted mask, deployed}             & 0.0463 & 0.0180 & 0.9365 & 0.9192 & 0.2011 \\
\midrule
Erode  16 px                                & 0.0465 & 0.0177 & 0.9363 & 0.9199 & 0.2041 \\
Erode  32 px                                & 0.0487 & 0.0186 & 0.9328 & 0.9143 & 0.2049 \\
Erode  64 px                                & 0.0528 & 0.0203 & 0.9287 & 0.9071 & 0.2055 \\
Dilate 16 px                                & 0.0467 & 0.0177 & 0.9372 & 0.9185 & 0.2011 \\
Dilate 32 px                                & 0.0494 & 0.0190 & 0.9338 & 0.9123 & 0.1989 \\
Dilate 64 px                                & 0.0522 & 0.0200 & 0.9305 & 0.9031 & 0.1977 \\
Shift  $x{=}32$ px                          & 0.0456 & 0.0171 & 0.9395 & 0.9240 & 0.2024 \\
Shift  $y{=}32$ px                          & 0.0470 & 0.0176 & 0.9397 & 0.9192 & 0.2012 \\
Shift  $xy{=}32$ px                         & 0.0474 & 0.0177 & 0.9386 & 0.9183 & 0.2009 \\
\bottomrule
\end{tabular}
\end{table}

\section{MaskPredictor training-time convergence}
\label{app:mask-convergence}
\begin{figure}[h]
  \centering
  \includegraphics[width=\linewidth]{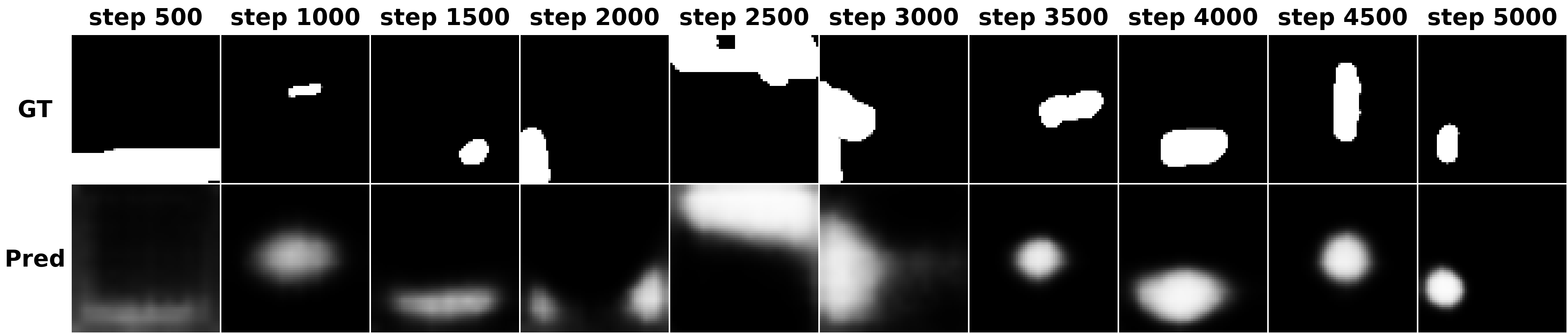}
  \caption{MaskPredictor convergence (variant G). Each column: a checkpoint step (500$\to$5000), GT mask (top) vs.\ predicted probability (bottom) at $64{\times}64$. By step 3000 predictions are compact and well-localized.}
  \label{fig:mask-convergence}
\end{figure}

\section{Qualitative results}
\label{app:qualitative}
\begin{figure}[h]
  \centering
  \includegraphics[width=\linewidth]{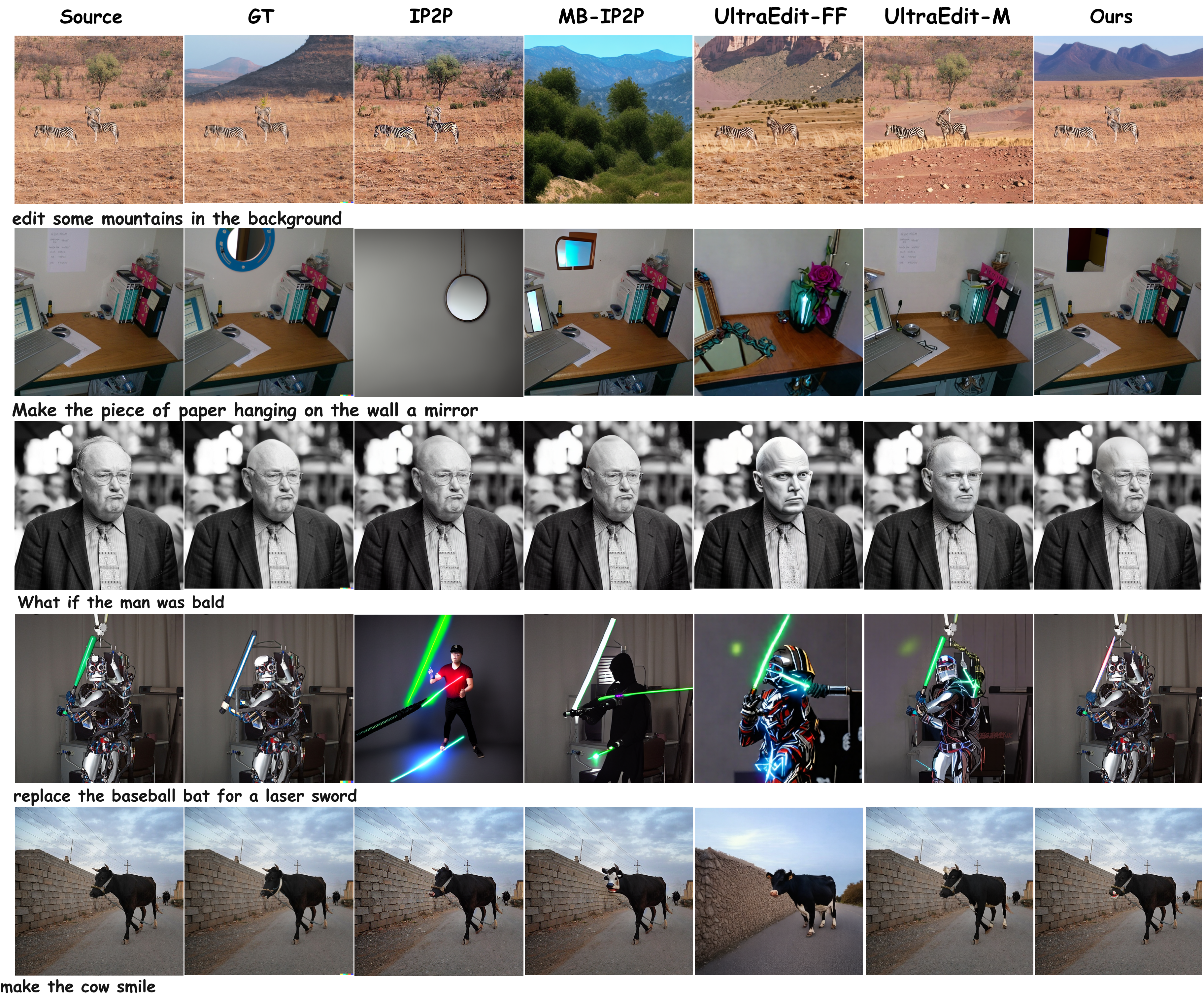}
  \caption{Qualitative comparison on MagicBrush dev. \method preserves unedited regions while applying the requested edit; baselines over-edit or miss the edit entirely.}
  \label{fig:qualitative}
\end{figure}

We show in Fig.~\ref{fig:qualitative} representative MagicBrush dev cases spanning the canonical edit categories: scene/background change (\emph{add mountains}), object replacement (\emph{paper $\to$ mirror}, \emph{baseball bat $\to$ laser sword}), attribute / expression change (\emph{make the cow smile}), and identity edit (\emph{the man is bald}). For each row we show source, GT, and four open-source baselines (InstructPix2Pix, MagicBrush, UltraEdit-FreeForm, UltraEdit-with-Mask) alongside \method (the deployed MaskPredictor configuration --- variant~G, no user mask at inference). Across all five rows \method preserves the unedited region nearly pixel-identically while applying the requested edit cleanly; baselines either over-edit the global appearance (IP2P, UltraEdit-FreeForm distort lighting / scene), miss the edit entirely (MagicBrush often falls back to near-identity), or hallucinate large structural changes (UltraEdit-with-Mask alters the surrounding scene).

\section{Block Adapter design choices}
\label{app:adapter-design}
\paragraph{Why a separate cross-attention rather than a joint-attention extension?}
A natural alternative to the per-block cross-attention adapter of \S\ref{sec:adapter} would be to concatenate the condition tokens to the existing joint-attention sequence and rely on the backbone's existing attention to absorb them. We deliberately avoided this for three reasons. First, modifying the input sequence length would invalidate the backbone's RoPE position encoding and require re-training of the rotary embeddings of the 60 blocks. Second, joint-attention scales as $O((L_\text{img}+L_\text{txt}+N_c)^2)$; injecting a separate cross-attention with $O(L_\text{img}\cdot N_c)$ keeps the cost linear in $N_c$ (we use $N_c{=}320$ in practice, of which 256 are mask tokens at $16{\times}16$ and 64 are instruction tokens). Third, the separate cross-attention path is ablatable: setting $g{\equiv}0$ recovers the un-adapted backbone exactly, which is the design that makes our variant A in Table~\ref{tab:ablation} a clean reference.

\paragraph{Source-token protection.}
Qwen-Image-Edit's joint-attention sequence interleaves \emph{target} image tokens (those being denoised) with \emph{source} reference tokens (the original image at zero noise, used as semantic context). Adding an adapter signal to the source tokens would corrupt the reference. We therefore mask out the gates of source tokens via a binary indicator $s_i\in\{0,1\}$ (provided by the dataloader): $g_i \leftarrow g_i\cdot(1-s_i)$. The combined effect of the binary $s_i$ and the soft SpatialGate $g_i$ (\S\ref{sec:spatialgate}) is that the adapter signal flows only into the target-branch tokens that lie inside the edit region --- a strong inductive bias that matches the human definition of ``local edit''. Empirically this single line accounts for a noticeable preservation gain on the dev split (cf.\ Table~\ref{tab:ablation}), by preventing the adapter from drifting the reference image semantics during long denoising trajectories.

\paragraph{Region-loss $\alpha$ sweep.}
We swept $\alpha\in\{1,2,4,8\}$ in early experiments and found that $\alpha=2$ struck the best edit-faithfulness vs.\ source-preservation trade-off on the dev split; larger $\alpha$ over-saturated the edit region and produced visible block-grid artefacts at the latent boundary, while $\alpha=1$ under-fit the edit region. We did not normalize away the $1$ in $w_i$ (i.e.\ kept a small but non-zero weight on the keep region) precisely so that the SpatialGate is still penalized when it leaks into the keep region.

\section{GEdit-Bench per-task breakdown}
\label{app:gedit-breakdown}
The GEdit-Bench-EN G\_O column is graded independently of G\_SC and G\_PQ by the GPT-4.1 judge and aggregates over a heterogeneous set of edit categories. Decomposing \method's per-category score makes the architecture's behaviour visible. \method is dominant on local edits with a well-defined region (\emph{text\_change} 8.53, \emph{subject-add} 8.28, \emph{color\_alter} 7.93, \emph{subject-replace} 7.63) --- exactly the segment the system targets. On global-edit categories that have no spatial extent for the SpatialGate to localize (\emph{motion\_change} 3.63, \emph{ps\_human} 3.29, \emph{style\_change} 4.92), the per-category score drops; this is the regime in which we recommend turning the adapter off via $g{\equiv}0$ and falling back to the un-adapted backbone (Appendix~\ref{app:discussion}).

\section{Discussion and Limitations}
\label{app:discussion}
\paragraph{When \method works best.}
Local edits with a clear semantic anchor benefit the most: object recolor, accessory swap, single-object remove, text rendering inside a region, and product-attribute change. On these we see the largest gains in CLIP-T and DINO and the largest reductions in L1 and L2 relative to the un-adapted backbone. This is exactly the segment that motivates the work and that current large joint-attention DiTs handle worst.

\paragraph{When \method should be turned off.}
The system's structural assumption is that the edit has a spatial extent that can be expressed as a (possibly soft) mask. There are three regimes where this assumption breaks down and the editor should fall back to the un-adapted backbone via $g{\equiv}0$:
(i) \emph{global stylization} (``make the photo warmer'', ``apply a vintage film look''): SpatialGate's edit region collapses to nearly the whole image and the gate signal stops carrying useful spatial information; pixel-preservation metrics in this regime are roughly equal to the baseline, but at the cost of additional adapter compute. (ii) \emph{Geometry-changing edits with no localized source region} (``put a hat on the dog'' where the dog has no hat): the GT mask covers the new content's location but there is no clean source signal to ``protect'' outside it; the protection is still useful, but the SpatialGate's role degrades to a uniform multiplier. (iii) \emph{Multi-object simultaneous edits} (``make the dress red and the wall green''): we currently produce a single union mask; per-object conditioning would be required to disentangle the two instructions.

\paragraph{Limitations.}
\begin{itemize}
  \item \textbf{MaskPredictor degrades on contextless edits.} On the local-edit setting that motivates this work the MaskPredictor is the recommended primary inference path (Table~\ref{tab:main}), but its accuracy drops sharply on instructions whose target region has no spatial extent (``make the photo warmer'', ``add some atmospheric haze''). In that regime the predicted mask collapses to a near-uniform near-1 sigmoid, and the user should fall back to the un-gated backbone via $g{\equiv}0$ rather than feed a noisy predicted mask through the SpatialGate.
  \item \textbf{Block-shared condition stream.} The Block Adapter shares the same condition stream $c$ across all 60 transformer blocks. Per-block conditioning would let early blocks attend to coarse layout and late blocks to fine texture, which is the dominant pattern in ControlNet-style branches. We chose the shared design for parameter efficiency: per-block conditioning would multiply the Condition Encoder cost by $60$ and offered only marginal gains in our preliminary experiments.
  \item \textbf{Single-mask assumption.} The current design assumes a single binary mask per edit. Multi-region edits (``recolor both shoes'' on a side-by-side composition) work in practice through a single union mask, but disentangled per-region instructions would need a per-region condition stream and a routing head; we leave this to future work.
  \item \textbf{Inherited backbone biases.} Like the backbone, \method inherits the upstream model's biases: lighting, ethnicity and product-category distributions of the training corpus of Qwen-Image-Edit. \method does not introduce new biases of its own (it only modulates spatially), but it also does not correct existing ones.
  \item \textbf{Training cost.} Although only $\sim$25\% of parameters are trainable, the forward pass still requires the full backbone, putting per-step memory at $\sim$70--78\,GB on bf16 with FSDP \texttt{HYBRID\_SHARD} and gradient checkpointing. We trained on $8\times$NVIDIA L20Y (80\,GB) GPUs; commodity 24\,GB cards cannot run the present system.
\end{itemize}

\paragraph{Broader impact.}
Improved local editing accelerates legitimate creative workflows (e-commerce, design, photo retouching) but also lowers the cost of producing visually plausible photo manipulations, including disinformation. Mitigations should pair the model with provenance tooling such as C2PA-style content credentials and watermarking baked into the diffusion decoder. We do not release the trained adapter checkpoints in this submission.

\section{Training data card}\label{app:datacard}

We give a qualitative description of the training corpus used for all variants
in Table~\ref{tab:main} and Table~\ref{tab:ablation}.

\paragraph{Instruction provenance.}
The MagicBrush half keeps the original human English instruction, translated once into Chinese; no per-sample rewriting. The internal text-edit half is rewritten by a Qwen3-class instruction-tuned language model from a short fixed template into a 30--80-character free-form edit prompt, conditioned on the (mask, source, target) triplet. The pre-rewrite template is preserved in a side field for auditability.

\paragraph{Mask provenance.}
No training masks come from segmentation models. The MagicBrush half uses the public human-annotated edit-region polygons. The text-edit half uses the ground-truth bounding region of the text rendered by the upstream pipeline --- i.e.\ the exact pixels modified by the rendering step --- not a segmentation prediction.

\paragraph{Task-type bias.}
The training shards do not carry an explicit per-row edit-type label (every row is generically tagged \texttt{Edit}). By source, the MagicBrush half spans the standard replace/remove/add/attribute taxonomy; the text-edit half is uniformly typography modification. Both halves are localized in the sense of having a single connected spatial extent at most resolutions, and the corpus contains essentially no global stylization, motion-change, or multi-object simultaneous-edit examples. We discuss the resulting failure modes in Appendix~\ref{app:discussion} (``When \method should be turned off'').

\paragraph{Preprocessing.}
Each row ships with offline-precomputed Qwen2.5-VL last-$L_h$ hidden states (Section~\ref{sec:condenc}); the VL encoder is therefore not loaded onto GPU during training, which is what makes the $\sim$70--78\,GB per-step memory budget on a 20B-parameter backbone feasible.

\end{document}